 \documentclass[tablecaption=bottom,wcp]{jmlr} 

\usepackage{booktabs}
\usepackage{aas_macros}
\usepackage[load-configurations=version-1]{siunitx} 

\usepackage{comment}

\theorembodyfont{\upshape}
\theoremheaderfont{\scshape}
\theorempostheader{:}
\theoremsep{\newline}

\jmlrproceedings{AABI 2019}{2nd Symposium on Advances in Approximate Bayesian Inference, 2019}

\title[Normalization Estimation]{Normalizing Constant Estimation with Gaussianized Bridge Sampling}

 \author{\Name{He Jia} \Email{he.jia.phy@gmail.com}\\
 \addr Department of Physics, Peking University, Beijing, 100871, China \\
 and \\ 
 Berkeley Center for Cosmological Physics, Department of Physics \\
 University of California, Berkeley, CA 94720, USA \\
 \AND
 \Name{Uro\v{s} Seljak} \Email{useljak@berkeley.edu}\\
 \addr  Department of Physics, Department of Astronomy \\
 University of California, Berkeley, CA 94720, USA \\
 and \\
 Lawrence Berkeley National Lab, 1 Cyclotron Road, Berkeley, CA 94720, USA \\
}

\begin{document}

\maketitle

\begin{abstract}
Normalizing constant (also called partition function, Bayesian evidence, or marginal likelihood) is one of the central goals of Bayesian inference, yet most of the existing methods are both expensive and inaccurate. Here we develop a new approach, starting from posterior samples obtained with a standard Markov Chain Monte Carlo (MCMC). We apply a novel Normalizing Flow (NF) approach to obtain an analytic density estimator from these samples, followed by Optimal Bridge Sampling (OBS) to obtain the normalizing constant. We compare our method which we call Gaussianized Bridge Sampling (GBS) to existing methods such as Nested Sampling (NS) and Annealed Importance Sampling (AIS) on several examples, showing our method is both significantly faster and substantially more accurate than these methods, and comes with a reliable error estimation.

\end{abstract}


\section{Introduction}

Normalizing constant, also called partition function, Bayesian evidence, 
or marginal likelihood, is the central object of Bayesian 
methodology. Despite its 
importance, existing methods are both inaccurate and slow, and may require specialized
tuning. One such method is Annealed 
Importance Sampling (AIS), and its alternative, Reverse AIS (RAIS), which can give stochastic lower and upper bounds to the normalizing constant, bracketing the true value \citep{neal2001annealed, grosse2015sandwiching}. However, as the tempered distribution may vary substantially with temperature, it can be expensive to obtain good samples at each temperature, which can lead to poor estimates \citep{murray2006nested}. 
Nested sampling (NS) is another popular alternative \citep{skilling2004nested, HandleyHobsonEtAl15}, 
which can be significantly more expensive than standard sampling methods in higher dimensions but, 
as we show, can also lead to very inaccurate estimates.
Moreover, there is no simple way to know how accurate the estimate is. 

Here we develop a new approach to the problem, combining 
Normalizing Flow (NF) density estimators with Optimal Bridge Sampling (OBS).
In a typical Bayesian inference application, we first obtain posterior samples using one of the standard Markov Chain Monte Carlo (MCMC) methods. In our approach we use these samples to derive the normalizing constant with 
relatively few additional likelihood evaluations required, making the additional 
cost of normalizing constant estimation small compared to posterior sampling. 
All of our calculations are run on standard CPU platforms,
and will be available in the \texttt{BayesFast} Python package.

\section{Bridge Sampling}

Let $p(\vec{x})$ and $q(\vec{x})$ be two possibly unnormalized distributions defined on $\Omega$, 
with normalizing constants $\mathcal{Z}_p$ and $\mathcal{Z}_q$. For any function
$\alpha(\vec{x})$ on $\Omega$, we have
\begin{equation}\label{eq:idendity}
\int_{\Omega} \alpha(\vec{x}) p(\vec{x}) q(\vec{x}) \text{d}\vec{x} = \mathcal{Z}_p \left< \alpha(\vec{x}) q(\vec{x}) \right>_p = \mathcal{Z}_q \left< \alpha(\vec{x}) p(\vec{x}) \right>_q,
\end{equation}
if the integral exists. Suppose that we have samples from both $p(\vec{x})$ and $q(\vec{x})$, 
and we know $\mathcal{Z}_q$, then \equationref{eq:idendity} gives
\begin{equation}\label{eq:bszp}
\mathcal{Z}_p= \frac{\left< \alpha(\vec{x}) p(\vec{x}) \right>_q}{\left< \alpha(\vec{x}) q(\vec{x}) \right>_p}\mathcal{Z}_q,
\end{equation}
which is the Bridge Sampling estimation of normalizing constant \citep{meng1996simulating}. 
It can be shown that many normalizing constant estimators, including Importance Sampling and Harmonic Mean, 
are special cases with different choices of bridge function $\alpha(\vec{x})$ \citep{gronau2017tutorial}.

For a given proposal function $q(\vec{x})$, an asymptotically optimal bridge function can be constructed, 
such that the ratio $r=\mathcal{Z}_p/\mathcal{Z}_q$ is given by the root of the following score function equation
\begin{equation}\label{eq:score}
S(r) = \sum_{i=1}^{n_p} \frac{n_q r q(\vec{x}_{p,i})}{n_p p(\vec{x}_{p,i}) + n_q r q(\vec{x}_{p,i})} - 
\sum_{i=1}^{n_q} \frac{n_p p(\vec{x}_{q,i})}{n_p p(\vec{x}_{q,i}) + n_q r q(\vec{x}_{q,i})}=0,
\end{equation}
where $n_p$ and $n_q$ are the numbers of samples from $p(\vec{x})$ and $q(\vec{x})$.
For $r\ge0$, $S(r)$ is monotonic and has a unique root,
so one can easily solve it with e.g. secant method. This estimator is \textit{optimal}, in the sense that its relative mean-square error is minimized \citep{chen2012monte}.

Choosing a suitable proposal $q(\vec{x})$ for Bridge Sampling can be challenging, as it requires a large overlap between $q(\vec{x})$ and $p(\vec{x})$. 
One approach is Warp Bridge Sampling (WBS) \citep{meng2002warp}, 
which transforms $p(\vec{x})$ to a Gaussian
with linear shifting, rescaling and symmetrizing. 
As we will show, this approach can be inaccurate or even fail completely for more complicated probability densities.

\section{Normalizing Flow Based Density Estimation}

As stated above, an appropriate proposal $q(\vec{x})$ which has large overlap with $p(\vec{x})$ is required for OBS to give accurate results. 
In a typical MCMC analysis we have samples from the posterior,
so one can obtain an approximate density estimation $q(\vec{x})$ from these samples 
using a bijective NF. 
In this approach one maps $p(\vec{x})$ to an unstructured 
distribution such as zero mean unit variance Gaussian $\mathcal{N}(0,\vec{I})$. 
For density evaluation we must also keep track of the Jacobian 
of transformation $|\text{d}\vec{\Psi}/\text{d}\vec{x}|$, so that our estimated distribution is $q(\vec{x})=\mathcal{N}(0,\vec{I})|\text{d}\vec{\Psi}/\text{d}\vec{x}|$, where $\vec{\Psi}(\vec{x})$ is the transformation. 
The probability density $q(\vec{x})$ is normalized, so we know $\mathcal{Z}_q=1$. 
There have been 
various methods of NF recently proposed in machine learning literature \citep{DinhKB14,DinhSB16,PapamakariosMP17}, which however failed on several examples we present below. Moreover, we observed that training
with these is very expensive and can easily dominate the 
overall computational cost. 

For these reasons we instead develop Iterative Neural Transform (INT), a new NF approach, details of which will be presented elsewhere.
It is based on combining optimal transport and information theory, 
repeatedly finding and transforming one dimensional marginals that are the most deviant
between the target and proposal (Gaussian) distributions.
After computing dual representation of Wasserstein-1 distance to find the maximally non-Gaussian directions, 
we apply a bijective transformation that maximizes the entropy along these directions.
For this we use a non-parametric spline based transformation that matches the 1-d cumulative distribution function (CDF) of the data to a Gaussian CDF, 
where kernel density estimation (KDE) is used to smooth the probability density marginals. 
We found that using a fixed number of 5 to 10 iterations is sufficient for evidence estimation, and the computational cost of our NF density estimation 
is small when compared to the cost of sampling.


\section{Proposed Method}

We propose the following Gaussianized Bridge Sampling (GBS) approach, which combines OBS with NF density estimation. In our typical application, 
we first run No-U-Turn Sampler (NUTS) \citep{hoffman2014no} to obtain $2n_p$ samples from $p(\vec{x})$ if its gradient is available, 
while affine invariant sampling \citep{foreman2013emcee} can be used in the gradient-free case.
To avoid underestimation of $\mathcal{Z}_p$ \citep{overstall2010default}, these $2n_p$ samples are divided into two batches, and we fit INT with the first batch of $n_p$ samples to obtain the proposal $q(\vec{x})$.
Then we draw $n_q$ samples from $q(\vec{x})$ and evaluate 
their corresponding $p(\vec{x})$, where $n_q$ is determined by an adaptive rule (see \appendixref{apd:sbe}).
We solve for the normalizing constant ratio $r$ with \equationref{eq:score}, using these $n_q$ samples from $q(\vec{x})$ and the second batch of $n_p$ samples from $p(\vec{x})$ (also evaluating their corresponding $q(\vec{x})$), and report the result in form of $\ln \mathcal{Z}_p$, with its 
error approximated by the relative mean-square error of $\mathcal{Z}_p$ given in \equationref{eq:error} \citep{chen2012monte}.

\section{Examples}

We used four test problems to compare the performance of various estimators.
See \appendixref{apd:examples} and \ref{apd:algorithms} for more details of the examples and algorithms.

(1) The 16-d \textit{Funnel} example is adapted from \cite{neal2003slice}. 
The funnel structure is common in Bayesian hierarchical models, 
and in practice it is recommended to reparameterize the model 
to overcome the pathology \citep{betancourt2015hamiltonian}. 
Here we stick to the original parameterization for test purpose.

(2) The 32-d \textit{Banana} example comes from a popular variant of multidimensional 
Rosenbrock function \citep{rosenbrock1960automatic}, 
which is composed of 16 uncorrelated 2-d bananas.
In addition, we apply a random 32-d rotation to the bananas, 
which makes all the parameters correlated with each other.

(3) The 48-d \textit{Cauchy} example is adapted from the \textit{LogGamma} example in \cite{feroz2013importance, buchner2016statistical}.
In contrast to the original example, where the mixture structure only exists 
in the first two dimensions, we place a mixture of two heavy-tailed Cauchy distributions along \textit{every} dimension.

(4) The 64-d \textit{Ring} example has strong non-linear correlation between the parameters, 
as the marginal distribution of every two successive parameters is ring-shaped.

\begin{figure}[htbp]
\floatconts
  {fig:results}
  {\caption{Comparison of normalizing constant estimators on the four examples, based on 64 simulations for each case. 
  Note that some panels use symmetrical logarithmic scale for x-axis. See the main text for
  abbreviation keys. 
  We show the quantiles of normalizing 
  constant results on the left and the number of total evaluations on the right, separately for likelihood and its gradient. 
  For WBS, GBS(L), GIS and GHM, the number of evaluations shown includes those required for posterior sampling, and the cost for evidence estimation alone is much smaller.}}
  {\vspace{-0.8cm}\includegraphics[width=\linewidth]{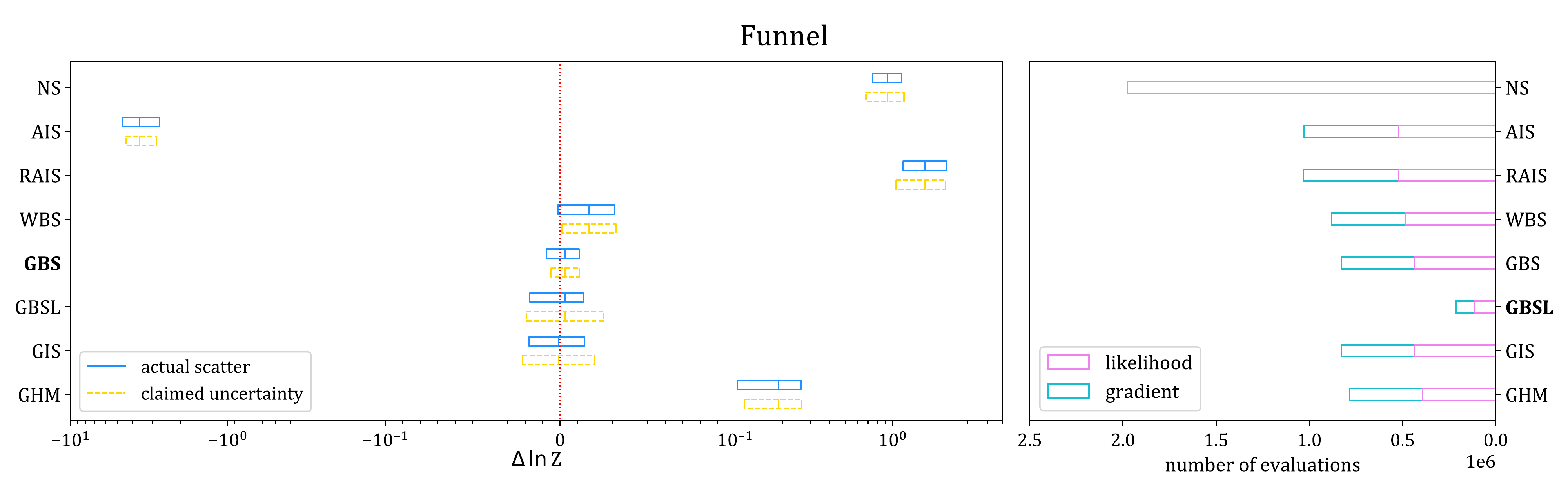}\vspace{-0.16cm},
  \includegraphics[width=\linewidth]{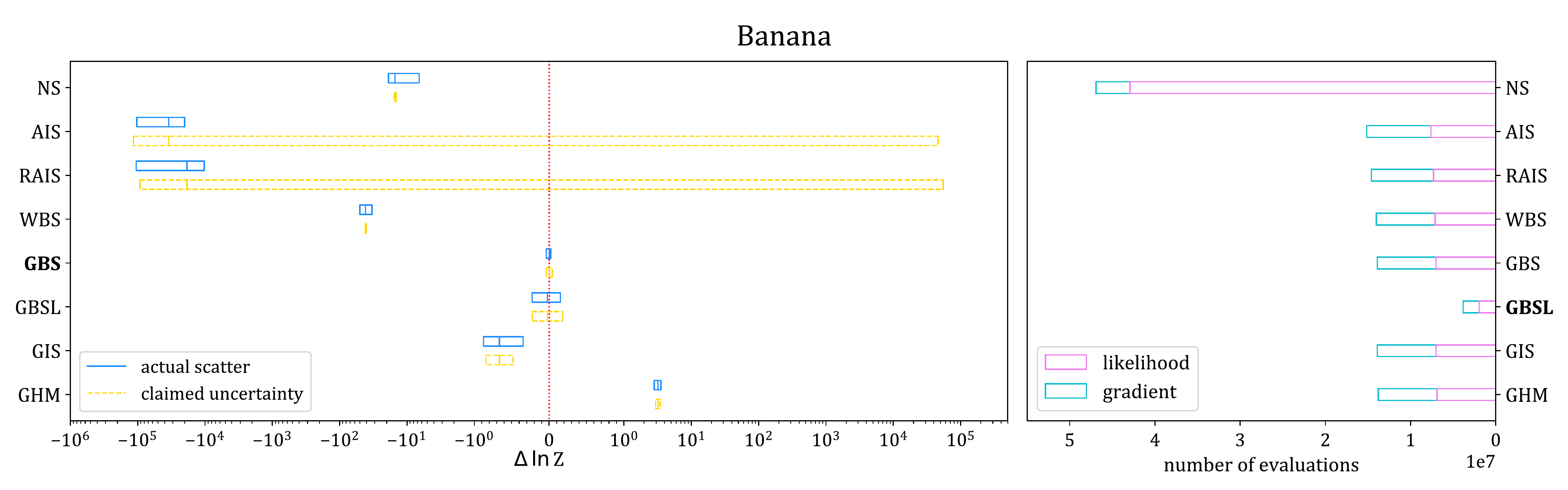}\vspace{-0.16cm},
  \includegraphics[width=\linewidth]{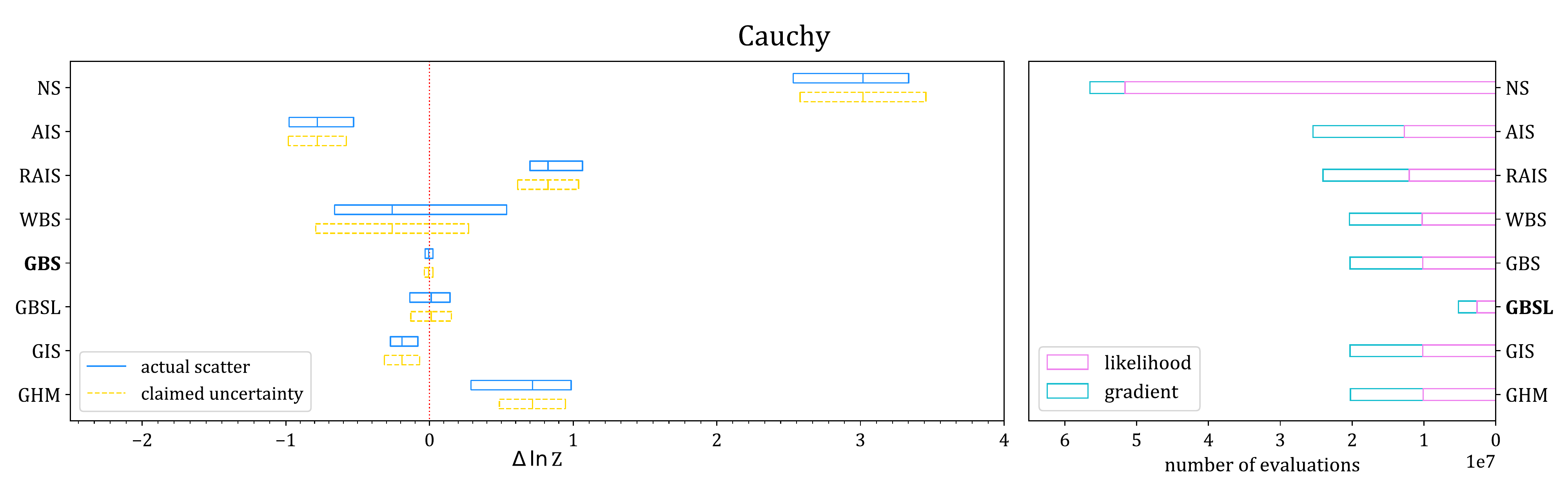}\vspace{-0.16cm},
  \includegraphics[width=\linewidth]{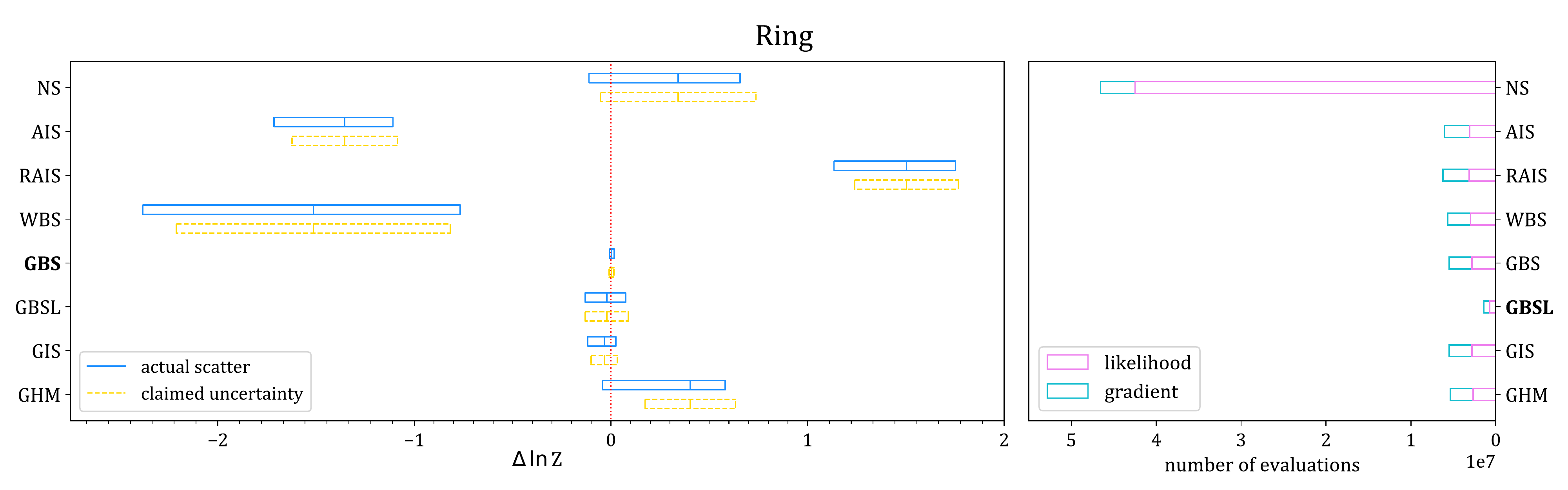}\vspace{-0.48cm}}
\end{figure}

See \figureref{fig:results} for a comparison of the estimators. For all of the four test examples, the proposed GBS algorithm gives the most accurate result and a valid error estimation. We use NS as implemented in \texttt{dynesty} \citep{speagle2019dynesty} with its default settings. 
For all other cases, we use NUTS as the MCMC transition operator. We chose to run (R)AIS with equal number of evaluations as our GBS, but as seen from \figureref{fig:results} this number is inadequate for (R)AIS, which needs about 10-100 times more evaluations to achieve sufficient accuracy (see \appendixref{apd:ais}). In contrast, if we run GBS with 4 times fewer evaluations (Gaussianized Bridge Sampling Lite, GBSL), we achieve an unbiased result with a larger error than GBS, but still smaller than other estimators. 
For comparison we also show results replacing  
OBS with IS (GIS) or HM (GHM), while 
still using INT for $q(\vec{x})$. Although GIS and GHM are 
better than NS or (R)AIS, they are worse than 
GBS(L), highlighting the importance of OBS. Finally, we also compare to WBS, 
which uses a very simple proposal distribution $q(\vec{x})$, and fails on several examples, 
highlighting the importance of using 
a more expressive NF for $q(\vec{x})$. 

For our GBS(L), most of evaluation time is used to get the posterior samples with standard MCMC, which is a typical Bayesian inference goal, and the additional cost to evaluate evidence is small compared to the MCMC (see \appendixref{apd:sbe}). In contrast, Thermodynamic Integration (TI) or (R)AIS is more expensive than posterior sampling, since the chains need to be accurate at \textit{every} intermediate state \citep{neal1993probabilistic}. The same 
comment applies to NS, which is more expensive than the MCMC approaches
we use here for posterior analysis, especially when non-informative prior is used.

\section{Conclusion}

We present a new method to estimate the normalizing constant (Bayesian evidence) in the 
context of Bayesian analysis. Our starting point are the samples from the posterior using 
standard MCMC based methods, and we assume that these have converged 
to the correct probability distribution. In our approach we combine OBS with INT, a novel
NF based density estimator, showing on several high dimensional 
examples that our method outperforms other approaches in terms of accuracy and 
computational cost, and provides a reliable error estimate.

\newpage

\bibliography{jmlr-sample,cosmo}

\begin{thebibliography}{24}
\providecommand{\natexlab}[1]{#1}
\providecommand{\url}[1]{\texttt{#1}}
\expandafter\ifx\csname urlstyle\endcsname\relax
  \providecommand{\doi}[1]{doi: #1}\else
  \providecommand{\doi}{doi: \begingroup \urlstyle{rm}\Url}\fi

\bibitem[Bennett(1976)]{bennett1976efficient}
Charles~H Bennett.
\newblock Efficient estimation of free energy differences from monte carlo
  data.
\newblock \emph{Journal of Computational Physics}, 22\penalty0 (2):\penalty0
  245--268, 1976.

\bibitem[Betancourt and Girolami(2015)]{betancourt2015hamiltonian}
Michael Betancourt and Mark Girolami.
\newblock Hamiltonian monte carlo for hierarchical models.
\newblock \emph{Current trends in Bayesian methodology with applications},
  79:\penalty0 30, 2015.

\bibitem[Buchner(2016)]{buchner2016statistical}
Johannes Buchner.
\newblock A statistical test for nested sampling algorithms.
\newblock \emph{Statistics and Computing}, 26\penalty0 (1-2):\penalty0
  383--392, 2016.

\bibitem[Chen et~al.(2012)Chen, Shao, and Ibrahim]{chen2012monte}
Ming-Hui Chen, Qi-Man Shao, and Joseph~G Ibrahim.
\newblock \emph{Monte Carlo methods in Bayesian computation}.
\newblock Springer Science \& Business Media, 2012.

\bibitem[Dinh et~al.(2014)Dinh, Krueger, and Bengio]{DinhKB14}
Laurent Dinh, David Krueger, and Yoshua Bengio.
\newblock {NICE:} non-linear independent components estimation.
\newblock \emph{CoRR}, abs/1410.8516, 2014.
\newblock URL \url{http://arxiv.org/abs/1410.8516}.

\bibitem[Dinh et~al.(2016)Dinh, Sohl{-}Dickstein, and Bengio]{DinhSB16}
Laurent Dinh, Jascha Sohl{-}Dickstein, and Samy Bengio.
\newblock Density estimation using real {NVP}.
\newblock \emph{CoRR}, abs/1605.08803, 2016.
\newblock URL \url{http://arxiv.org/abs/1605.08803}.

\bibitem[Feroz et~al.(2013)Feroz, Hobson, Cameron, and
  Pettitt]{feroz2013importance}
F~Feroz, MP~Hobson, E~Cameron, and AN~Pettitt.
\newblock Importance nested sampling and the multinest algorithm.
\newblock \emph{arXiv preprint arXiv:1306.2144}, 2013.

\bibitem[Foreman-Mackey et~al.(2013)Foreman-Mackey, Hogg, Lang, and
  Goodman]{foreman2013emcee}
Daniel Foreman-Mackey, David~W Hogg, Dustin Lang, and Jonathan Goodman.
\newblock emcee: the mcmc hammer.
\newblock \emph{Publications of the Astronomical Society of the Pacific},
  125\penalty0 (925):\penalty0 306, 2013.

\bibitem[Fr{\"u}hwirth-Schnatter(2004)]{fruhwirth2004estimating}
Sylvia Fr{\"u}hwirth-Schnatter.
\newblock Estimating marginal likelihoods for mixture and markov switching
  models using bridge sampling techniques.
\newblock \emph{The Econometrics Journal}, 7\penalty0 (1):\penalty0 143--167,
  2004.

\bibitem[Gronau et~al.(2017)Gronau, Sarafoglou, Matzke, Ly, Boehm, Marsman,
  Leslie, Forster, Wagenmakers, and Steingroever]{gronau2017tutorial}
Quentin~F Gronau, Alexandra Sarafoglou, Dora Matzke, Alexander Ly, Udo Boehm,
  Maarten Marsman, David~S Leslie, Jonathan~J Forster, Eric-Jan Wagenmakers,
  and Helen Steingroever.
\newblock A tutorial on bridge sampling.
\newblock \emph{Journal of mathematical psychology}, 81:\penalty0 80--97, 2017.

\bibitem[Grosse et~al.(2015)Grosse, Ghahramani, and
  Adams]{grosse2015sandwiching}
Roger~B Grosse, Zoubin Ghahramani, and Ryan~P Adams.
\newblock Sandwiching the marginal likelihood using bidirectional monte carlo.
\newblock \emph{arXiv preprint arXiv:1511.02543}, 2015.

\bibitem[{Handley} et~al.(2015){Handley}, {Hobson}, and
  {Lasenby}]{HandleyHobsonEtAl15}
W.~J. {Handley}, M.~P. {Hobson}, and A.~N. {Lasenby}.
\newblock {POLYCHORD: next-generation nested sampling}.
\newblock \emph{\mnras}, 453:\penalty0 4384--4398, November 2015.
\newblock \doi{10.1093/mnras/stv1911}.

\bibitem[Hoffman and Gelman(2014)]{hoffman2014no}
Matthew~D Hoffman and Andrew Gelman.
\newblock The no-u-turn sampler: adaptively setting path lengths in hamiltonian
  monte carlo.
\newblock \emph{Journal of Machine Learning Research}, 15\penalty0
  (1):\penalty0 1593--1623, 2014.

\bibitem[Meng and Schilling(2002)]{meng2002warp}
Xiao-Li Meng and Stephen Schilling.
\newblock Warp bridge sampling.
\newblock \emph{Journal of Computational and Graphical Statistics}, 11\penalty0
  (3):\penalty0 552--586, 2002.

\bibitem[Meng and Wong(1996)]{meng1996simulating}
Xiao-Li Meng and Wing~Hung Wong.
\newblock Simulating ratios of normalizing constants via a simple identity: a
  theoretical exploration.
\newblock \emph{Statistica Sinica}, pages 831--860, 1996.

\bibitem[Murray et~al.(2006)Murray, MacKay, Ghahramani, and
  Skilling]{murray2006nested}
Iain Murray, David MacKay, Zoubin Ghahramani, and John Skilling.
\newblock Nested sampling for potts models.
\newblock In \emph{Advances in Neural Information Processing Systems}, pages
  947--954, 2006.

\bibitem[Neal(1993)]{neal1993probabilistic}
Radford~M Neal.
\newblock \emph{Probabilistic inference using Markov chain Monte Carlo
  methods}.
\newblock Department of Computer Science, University of Toronto Toronto,
  Ontario, Canada, 1993.

\bibitem[Neal(2001)]{neal2001annealed}
Radford~M Neal.
\newblock Annealed importance sampling.
\newblock \emph{Statistics and computing}, 11\penalty0 (2):\penalty0 125--139,
  2001.

\bibitem[Neal et~al.(2003)]{neal2003slice}
Radford~M Neal et~al.
\newblock Slice sampling.
\newblock \emph{The annals of statistics}, 31\penalty0 (3):\penalty0 705--767,
  2003.

\bibitem[Overstall and Forster(2010)]{overstall2010default}
Antony~M Overstall and Jonathan~J Forster.
\newblock Default bayesian model determination methods for generalised linear
  mixed models.
\newblock \emph{Computational Statistics \& Data Analysis}, 54\penalty0
  (12):\penalty0 3269--3288, 2010.

\bibitem[Papamakarios et~al.(2017)Papamakarios, Murray, and
  Pavlakou]{PapamakariosMP17}
George Papamakarios, Iain Murray, and Theo Pavlakou.
\newblock Masked autoregressive flow for density estimation.
\newblock In \emph{Advances in Neural Information Processing Systems 30: Annual
  Conference on Neural Information Processing Systems 2017, 4-9 December 2017,
  Long Beach, CA, {USA}}, pages 2335--2344, 2017.
\newblock URL
  \url{http://papers.nips.cc/paper/6828-masked-autoregressive-flow-for-density-estimation}.

\bibitem[Rosenbrock(1960)]{rosenbrock1960automatic}
HoHo Rosenbrock.
\newblock An automatic method for finding the greatest or least value of a
  function.
\newblock \emph{The Computer Journal}, 3\penalty0 (3):\penalty0 175--184, 1960.

\bibitem[Skilling(2004)]{skilling2004nested}
John Skilling.
\newblock Nested sampling.
\newblock In \emph{AIP Conference Proceedings}, volume 735, pages 395--405.
  AIP, 2004.

\bibitem[Speagle(2019)]{speagle2019dynesty}
Joshua~S Speagle.
\newblock dynesty: A dynamic nested sampling package for estimating bayesian
  posteriors and evidences.
\newblock \emph{arXiv preprint arXiv:1904.02180}, 2019.

\end{thebibliography}

\newpage

\appendix

\section{Details of Examples}\label{apd:examples}

\subsection{16-d Funnel}

The model likelihood is
\begin{equation}\label{eq:funnel}
\mathcal{L}=\mathcal{N}(x_1\,|\,0, a^2) \prod_{i=2}^n \mathcal{N}(x_i\,|\,0, \exp(2bx_1)),
\quad a=1,
\quad b=0.5,
\quad n=16,
\end{equation}
with flat prior $x_1\sim\mathcal{U}(-4,4)$, $x_{2:n}\sim\mathcal{U}(-30,30)$. We use $\ln\mathcal{Z}_p=-63.4988$ as the fiducial value, and the corner plot of the first four dimensions is shown in \figureref{fig:funnel}.

\begin{figure}[htbp]
\floatconts
  {fig:funnel}
  {\caption{Corner plot for the \textit{Funnel} example.}}
  {\includegraphics[width=0.64\linewidth]{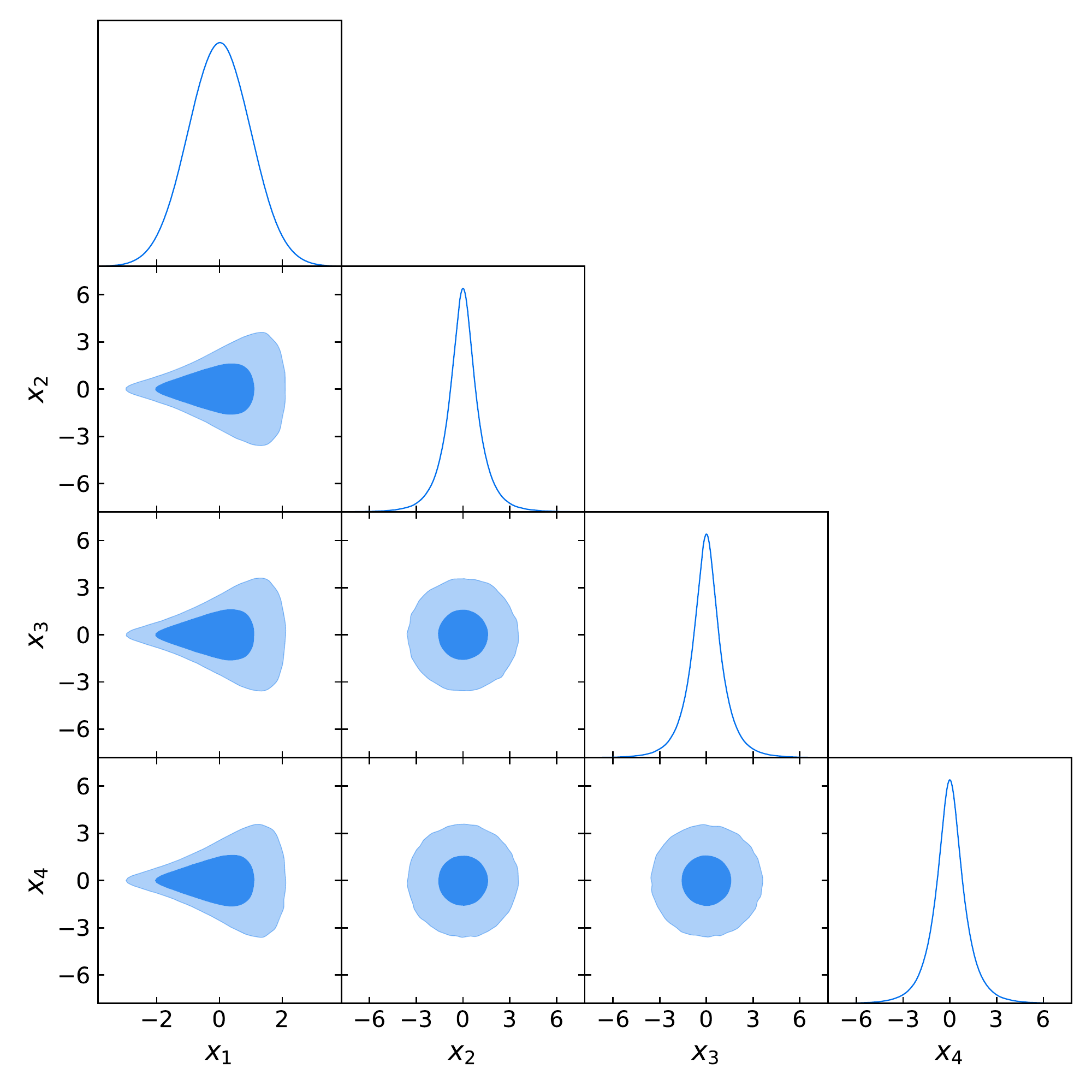}}
\end{figure}

\subsection{32-d Banana}

\begin{figure}[htbp]
\floatconts
  {fig:banana}
  {\caption{Corner plot for the \textit{Banana} example. Top: without random rotation. Bottom: with random rotation.}}
  {\includegraphics[width=0.64\linewidth]{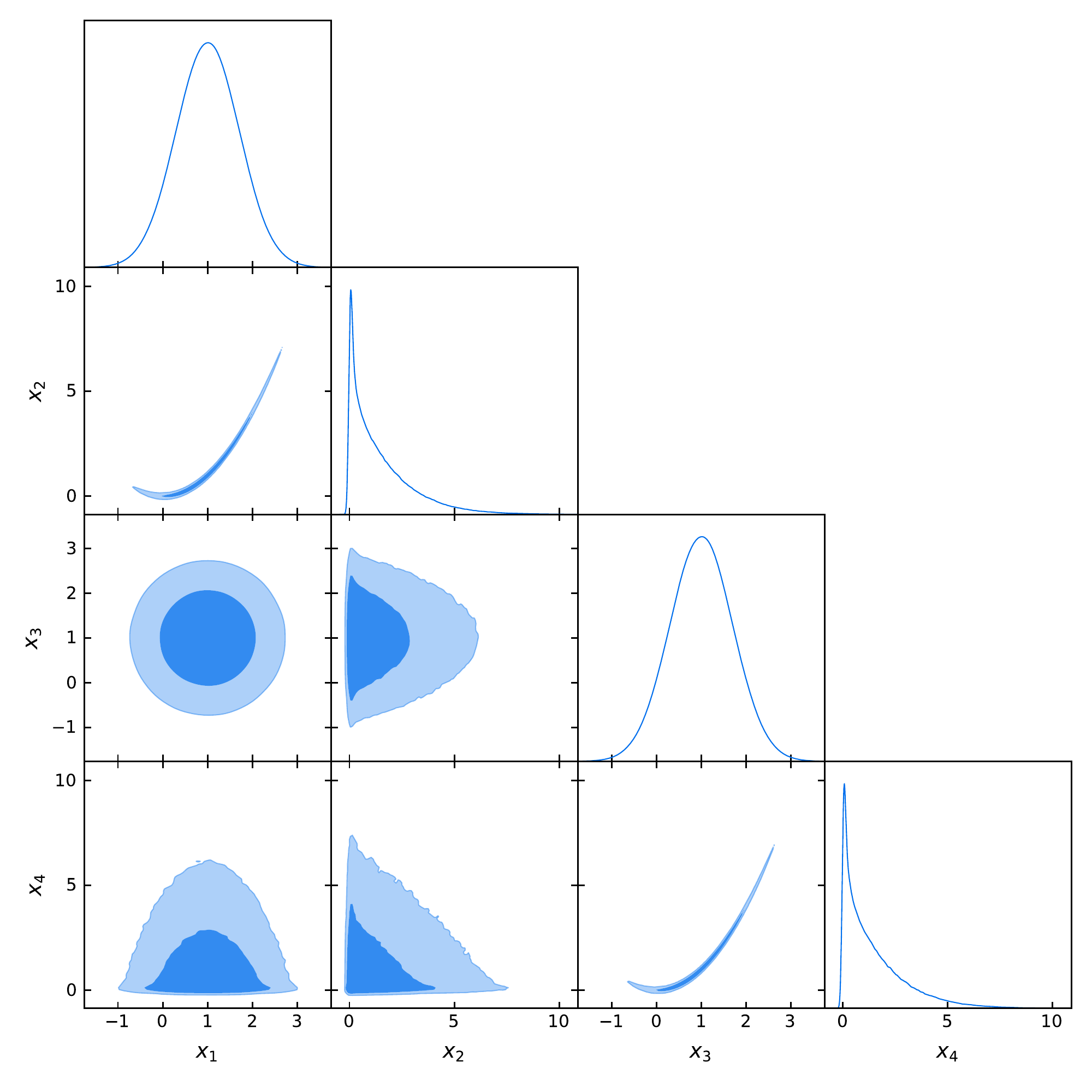}
  \includegraphics[width=0.64\linewidth]{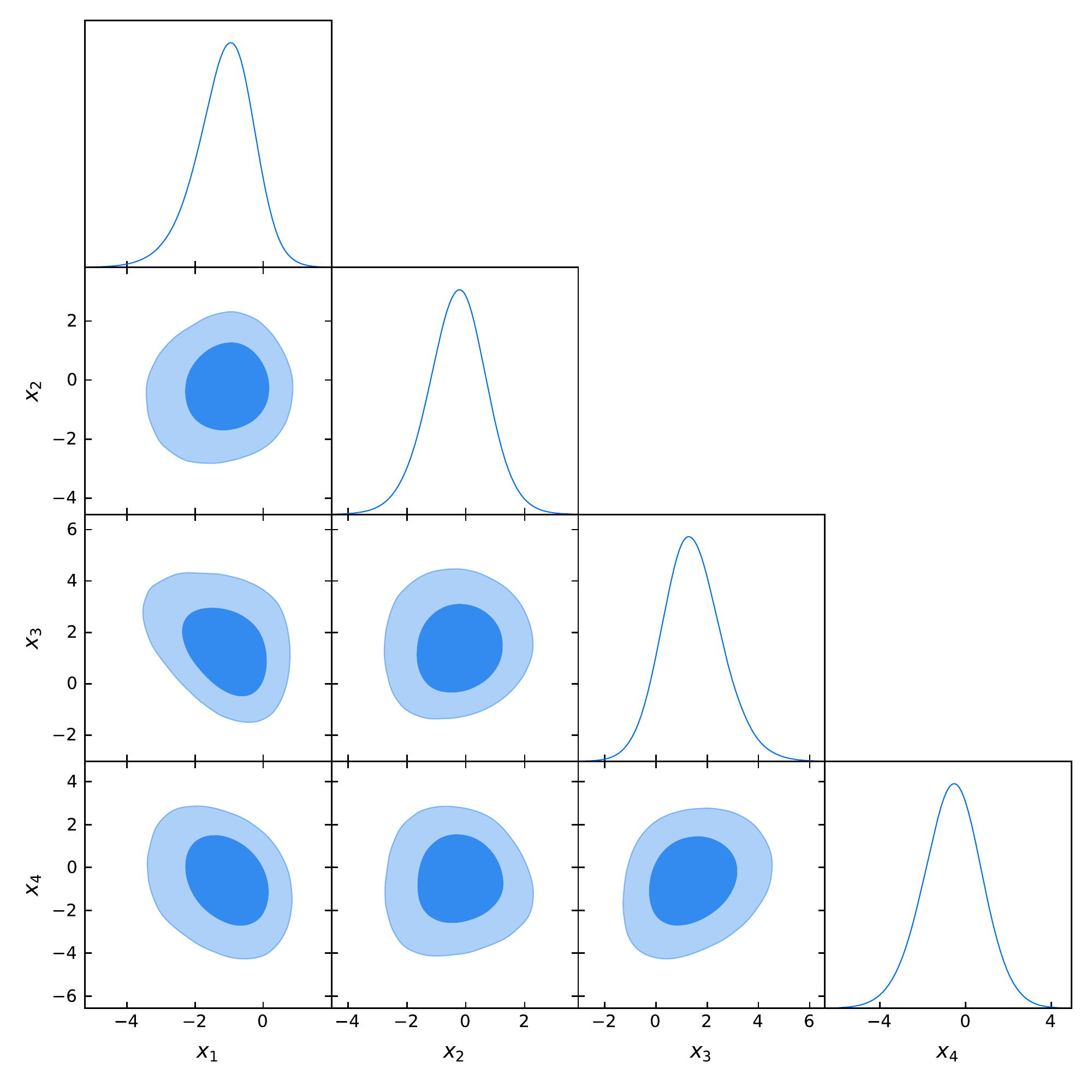}}
\end{figure}

The model likelihood is
\begin{equation}\label{eq:banana}
\ln\mathcal{L}=-\sum_{i=1}^{n/2}\left[\,(y_{2i-1}^2-y_{2i})^2/Q+(y_{2i-1}-1)^2\,\right],
\quad \vec{y}=\vec{A}\vec{x},
\quad Q=0.01,
\quad n=32,
\end{equation}
with flat prior $\mathcal{U}(-15,15)$ on all the parameters. The rotation matrix $\vec{A}$ is generated  from a random sample of $\text{SO}(n)$, and the same $\vec{A}$ is used for all the simulations. We use $\ln\mathcal{Z}_p=-127.364$ as the fiducial value, and the corner plot of the first four dimensions, without or with the random rotation, is shown in \figureref{fig:banana}. The strong degeneracy can no longer be identified in the plot once we apply the rotation, however it still exists and hinders most estimators from getting reasonable results.

\subsection{48-d Cauchy}

The model likelihood is
\begin{equation}\label{eq:cauchy}
\mathcal{L}=\prod_{i=1}^n \frac{1}{2}\left[\,\text{Cauchy}(x_i|\mu,\sigma) + 
\text{Cauchy}(x_i|{-\mu},\sigma)\,\right],
\quad \mu=5,
\quad \sigma=1,
\quad n=48,
\end{equation}
with flat prior $\mathcal{U}(-100,100)$ on all the parameters. We use $\ln\mathcal{Z}_p=-254.627$ as the fiducial value, and the corner plot of the first four dimensions is shown in \figureref{fig:cauchy}.

\begin{figure}[htbp]
\floatconts
  {fig:cauchy}
  {\caption{Corner plot for the \textit{Cauchy} example.}}
  {\includegraphics[width=0.64\linewidth]{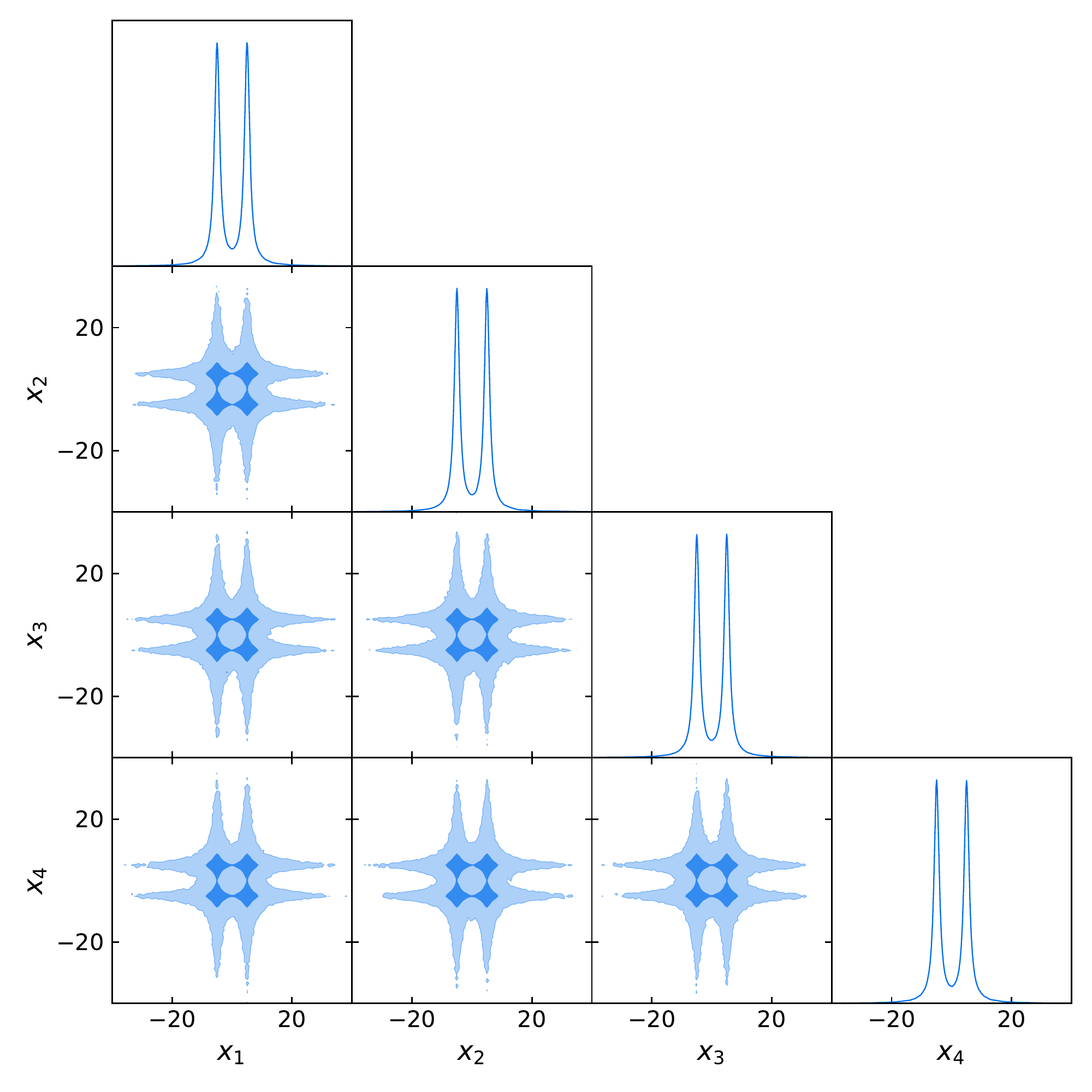}}
\end{figure}

\subsection{64-d Ring}

The model likelihood is
\begin{equation}\label{eq:ring}
\ln\mathcal{L}=-\left[\frac{(x_n^2+x_1^2-a)^2}{b}\right]^2
-\sum_{i=1}^{n-1}\left[\frac{(x_i^2+x_{i+1}^2-a)^2}{b}\right]^2,
\quad a=2,
\quad b=1,
\quad n=64,
\end{equation}
with flat prior $\mathcal{U}(-5,5)$ on all the parameters. We use $\ln\mathcal{Z}_p=-114.492$ as the fiducial value, and the corner plot of the first four dimensions is shown in \figureref{fig:ring}.

\begin{figure}[htbp]
\floatconts
  {fig:ring}
  {\caption{Corner plot for the \textit{Ring} example.}}
  {\includegraphics[width=0.64\linewidth]{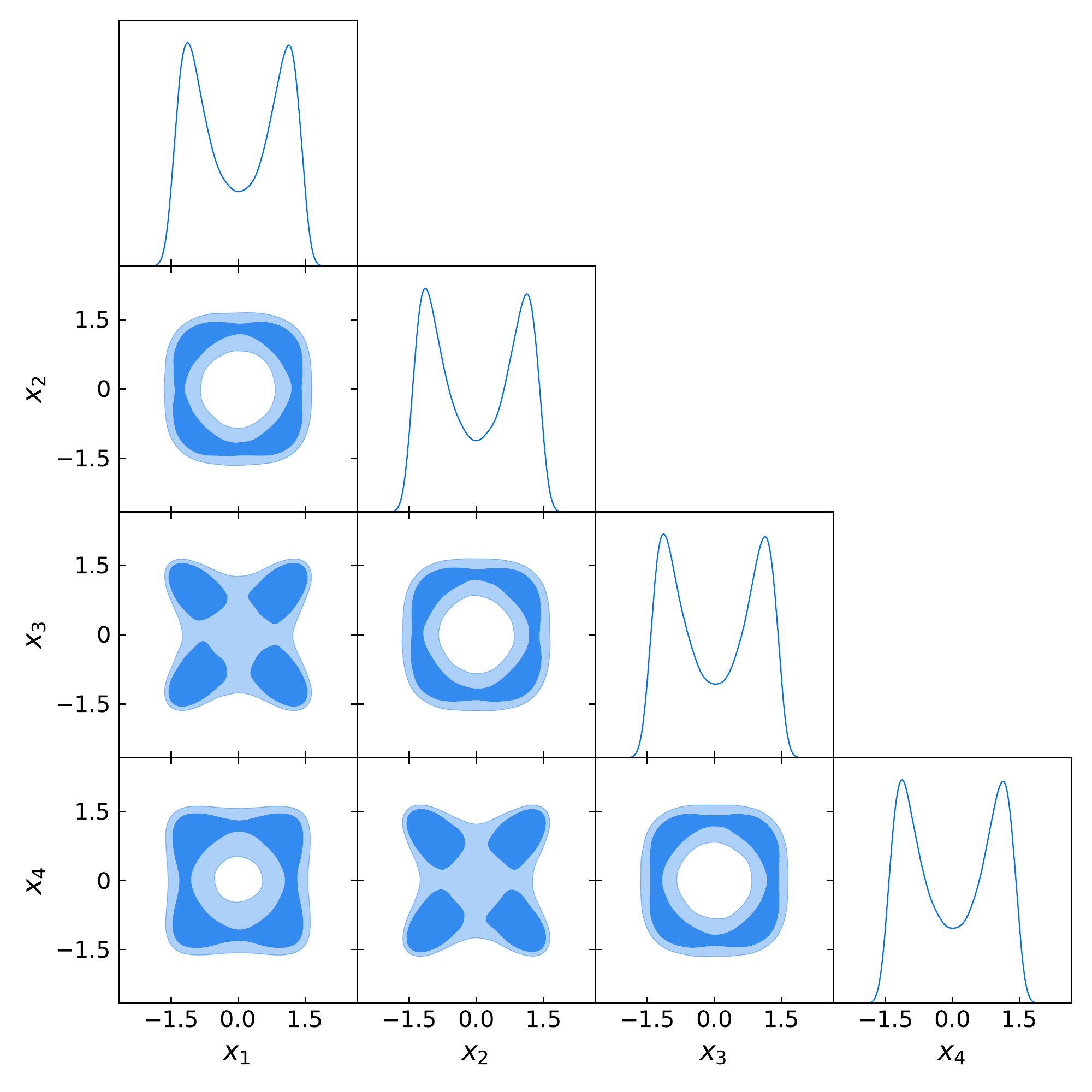}}
\end{figure}

\section{Details of Algorithms}\label{apd:algorithms}

\subsection{Obtaining Fiducial Values}

Analytic normalizing constants for the (unconstrained) likelihood of the \textit{Funnel}, \textit{Banana} and \textit{Cauchy} examples are available.
For the \textit{Funnel} and \textit{Banana} examples, 
we account for the effect of finite prior range by simply generating a large amount of samples and counting the fraction of the samples that are inside the prior range. 
For the \textit{Cauchy} example, we directly evaluate its CDF.
For the \textit{Ring} example, the fiducial normalizing constant is obtained by a long AIS and RAIS run, with 300,000 intermediate states and 64 chains in both directions.

\subsection{Nested Sampling}

We use dynamic NS implemented in \texttt{dynesty}, which is considered more efficient than static NS.
Traditionally, NS does not need the gradient of the likelihood, at the cost of lower sampling efficiency in high dimensions.
Since analytic gradient of the four examples is available, we follow \texttt{dynesty}'s default setting, which requires the gradient to perform Hamitonian Slice Sampling for dimensions $d>20$.
While for dimensions $10 \leq d \leq 20$, random walks sampling is used instead.
\texttt{dynesty} also provides an error estimate for the evidence; see \cite{speagle2019dynesty} for details.

\subsection{(Reversed) Annealed Importance Sampling}\label{apd:ais}

For (R)AIS, we divide the warm-up iterations of NUTS into two equal stages, and the (flat) prior is used as the base density. In the first stage, we set $\beta=0.5$ and adapt the mass matrix and step size of NUTS, which acts as a compromise between the possibly broad prior and narrow posterior. In the second stage, we set $\beta=0$ ($\beta=1$) for AIS (RAIS) to get samples from the prior (posterior). After warm-up, we use the following sigmoidal schedule to perform annealing,
\begin{equation}\label{eq:schedule}
\tilde{\beta}_t=\sigma\left(\delta\left(\frac{2t}{T-1}-1\right)\right),\qquad
\beta_t=\frac{\tilde{\beta}_t-\tilde{\beta}_0}{\tilde{\beta}_{T-1}-\tilde{\beta}_0},\qquad
0\leq t\leq T-1,
\end{equation}
where $\sigma$ denotes the logistic sigmoid function and we set $\delta=4$ \citep{grosse2015sandwiching}. We use 1,000 warm-up iterations for all the four examples, and adjust the number of states $T$ so that it needs roughly the same number of evaluations
as GBS in total. The exact numbers are listed in \tableref{tab:ais}. 
We run 16 chains for each case, 
and average reported $\ln\mathcal{Z}_p$ of different chains, 
which gives a stochastic lower (upper) bound for AIS (RAIS) 
according to Jensen's inequality.
The uncertainty is estimated from the scatter of different chains,
and should be understood as the error of the lower (upper) bound of $\ln\mathcal{Z}_p$,
instead of $\ln\mathcal{Z}_p$ itself.

\begin{table}[hbtp]
\floatconts
  {tab:ais}
  {\caption{The number of states $T$ used by (R)AIS.}}
  {\begin{tabular}{lllll}
  \toprule
   & Funnel & Banana & Cauchy & Ring\\
  \midrule
  AIS & 800 & 2000 & 3000 & 3500 \\
  RAIS & 700 & 1500 & 2500 & 3000 \\
  \bottomrule
  \end{tabular}}
\end{table}

Using the mass matrix and step size of NUTS adapted at 
$\beta=0.5$, and the prior as base density, may account for the phenomenon that RAIS failed to give an upper bound in the \textit{Banana} example: the density is very broad at high temperatures and very narrow at low temperatures, which is difficult for samplers adapted at a single $\beta$. One may remedy this issue by using a better base density that is closer to the posterior, but this will require delicate hand-tuning and is beyond the scope of this paper.
While the upper (lower) bounds of (R)AIS 
are valid in the limit of a very large 
number of samples, achieving this limit may be 
extremely costly in practice. 

\subsection{Sample-Based Estimators}\label{apd:sbe}

The remaining normalizing constant estimators require a sufficient number of samples from $p(\vec{x})$, which we obtain with NUTS.
For WBS, GBS, GIS and GHM, we run 8 chains with 2,500 iterations for the \textit{Funnel} and \textit{Banana} examples, and 5,000 iterations for the \textit{Cauchy} and \textit{Ring} examples, including the first 20\% warm-up iterations, which are removed from the samples. Then we fit INT using 10 iterations for GBS, GIS and GHM, whose computation cost (a few seconds for the \textit{Funnel} example) is small or negligible relative to NUTS sampling, and does not depend on the cost of $\ln p(\vec{x})$ evaluations. For GBSL, the number of NUTS chains, NUTS iterations and INT iterations are all reduced by half, leading to a factor of four decrease in the total computation cost. 

The relative mean-square error of OBS is minimized and
given by
\begin{equation}\label{eq:error}
\widehat{RE^2_{\text{OBS}}} = \frac{1}{n_q} \frac{\text{Var}_q(f_1(\vec{x}))}{\text{E}_q^2(f_1(\vec{x}))} + \frac{\tau_{f_2}}{n_p} \frac{\text{Var}_p(f_2(\vec{x}))}{\text{E}_p^2(f_2(\vec{x}))},
\end{equation}
where $f_1(\vec{x})=\frac{p'(\vec{x})}{s_p p'(\vec{x}) + s_q q(\vec{x})}$,
$f_2(\vec{x})=\frac{q(\vec{x})}{s_p p'(\vec{x}) + s_q q(\vec{x})}$,
$s_p=\frac{n_p}{n_p+n_q}$, $s_q=\frac{n_q}{n_p+n_q}$. 
Here $p'(\vec{x})=p(\vec{x})/\mathcal{Z}_p$ and $q(\vec{x})$ should be normalized densities.
We assume the samples from $q(\vec{x})$ are independent, whereas the samples from $p(\vec{x})$ may be autocorrelated, and $\tau_{f_2}$ is the integrated autocorrelation time of $f_2(\vec{x}_p)$ \citep{fruhwirth2004estimating},
which is estimated by the \texttt{autocorr} module in \texttt{emcee} \citep{foreman2013emcee}.
Analogous expressions can be derived for IS and HM,
\begin{gather}\label{eq:error2}
\widehat{RE^2_{\text{IS}}} = \frac{1}{n_q} \text{Var}_q(f_{\text{IS}}(\vec{x})),\qquad
f_{\text{IS}}(\vec{x}) = \frac{p'(\vec{x})}{q(\vec{x})},\nonumber\\
\widehat{RE^2_{\text{HM}}} = \frac{\tau_{f_{\text{HM}}}}{n_p} \text{Var}_p(f_{\text{HM}}(\vec{x})),\qquad
f_{\text{HM}}(\vec{x}) = \frac{q(\vec{x})}{p'(\vec{x})}.
\end{gather}
The claimed uncertainty in \figureref{fig:results} is obtained by assuming that the error is Gaussian distributed.

There can be different strategies to allocate samples for BS. 
In the literature, it is recommended that one draws samples from $p(\vec{x})$ and $q(\vec{x})$ based on equal-sample-size or equal-time allocation \citep{bennett1976efficient, meng1996simulating}.
Since NUTS based sampling usually requires at least hundreds of evaluations to obtain one effective sample from $p(\vec{x})$ in high dimensions \citep{hoffman2014no}, which is orders of magnitude more expensive than our NF based sampling for $q(\vec{x})$, it could be advantageous to set $n_q>n_p$. 
Throughout this paper, the following adaptive strategy is adopted to determine $n_q$ for GBS(L). 
After obtaining $2n_p$ samples from $p(\vec{x})$, we divide them into two equal batches, which will be used for fitting the proposal $q(\vec{x})$ and evaluating the evidence, respectively.
As an starting point, we draw $n_{q,0}=n_p$ samples from $q(\vec{x})$ and estimate the error of OBS using \equationref{eq:error}.
Note that the right side of \equationref{eq:error} is composed of two terms, and only the first term will decrease as one increases $n_q$ but fixes $n_p$. 
Assuming that current samples provide an accurate estimate of the variance and expectation terms in \equationref{eq:error}, one can solve for $n_q$ such that $f_{\text{err}}$, the fraction of $q(\vec{x})$ contributions in \equationref{eq:error}, is equal to some specified value, which we set to 0.1. 
Since the $n_{q,0}$ samples from $q(\vec{x})$ can be reused, if $n_q<n_{q,0}$, no additional samples are required and we set $n_q=n_{q,0}$.
On the other hand, we also require that $f_{\text{eva}}$, the fraction of $p(\vec{x})$ evaluations that are used for the $q(\vec{x})$ samples, is no larger than 0.1, although this constraint is usually not activated in practice.

We use 0.1 as the default values of $f_{\text{err}}$ and $f_{\text{eva}}$, so that the additional cost of evidence evaluation is small relative to the cost of sampling, while using a larger $n_q$ alone can no longer significantly improve the accuracy of normalizing constant. 
However, if one wants to put more emphasis on posterior sampling (evidence estimation), a larger (smaller) $f_{\text{err}}$ and/or smaller (larger) $f_{\text{eva}}$ can be used.
In principle, it is also possible to use different number of $p(\vec{x})$ samples to fit the proposal and evaluate the evidence, in contrast to equal split used in \cite{overstall2010default}, which we leave for feature research.

For GIS and WBS, we use the same $n_q$ as solved for GBS(L).
No samples from $p(\vec{x})$ are required to estimate normalizing constant for GIS, so in this case all the $2n_p$ samples will be used to fit INT.
While for GHM, no samples from $q(\vec{x})$ are required.
Note that for WBS, the additional $p(\vec{x})$ evaluations required for evidence estimation is $n_p+2n_q$ instead of $n_q$, which comes from the symmetrization of $\ln p(\vec{x})$.

\end{document}